\documentclass[letterpaper]{article} 
\usepackage[draft]{aaai25}  
\usepackage{times}  
\usepackage{helvet}  
\usepackage{courier}  
\usepackage[hyphens]{url}  
\usepackage{graphicx} 
\usepackage{natbib}  
\usepackage{caption} 
\usepackage{algorithm}
\usepackage{algorithmic}

\usepackage{tikz}

\usepackage{enumitem}
\usepackage{multirow}
\usepackage{multicol}
\usepackage{subcaption}
\usepackage{booktabs}
\usepackage{amsmath}

\usepackage{newfloat}
\usepackage{listings}
\DeclareCaptionStyle{ruled}{labelfont=normalfont,labelsep=colon,strut=off} 
\floatstyle{ruled}
\newfloat{listing}{tb}{lst}{}
\floatname{listing}{Listing}
\newcommand{\noteb}[1]{{}}

\usepackage{makecell}

\title{Bits-to-Photon: End-to-End Learned Scalable Point Cloud Compression for Direct Rendering}
\author{
    Yueyu Hu, Ran Gong, Yao Wang
}
\affiliations{
     Department of Electrical and Computer Engineering \\
    Tandon School of Engineering, New York University\\

    370 Jay St, Brooklyn, NY 11201\\
    \{yyhu, rg4827, yaowang\}@nyu.edu
}

\usepackage{bibentry}

\begin{document}

\maketitle

\begin{abstract}
    Point cloud is a promising 3D representation for volumetric streaming in emerging AR/VR applications. Despite recent advances in point cloud compression, decoding and rendering high-quality images from lossy compressed point clouds is still challenging in terms of quality and complexity, making it a major roadblock to achieve real-time 6-Degree-of-Freedom video streaming. In this paper, we address this problem by developing a point cloud compression scheme that generates a bit stream that can be directly decoded to renderable 3D Gaussians. The encoder and decoder are jointly optimized to consider both bit-rates and rendering quality. It   significantly improves the rendering quality while substantially reducing decoding and rendering time, compared to existing point cloud compression methods. Furthermore, the proposed scheme generates a scalable bit stream, allowing multiple levels of details at different bit-rate ranges. Our method supports real-time color decoding and rendering of high quality point clouds, thus paving the way  for interactive 3D streaming applications with free view points.
\end{abstract}

%

\section{Introduction}
\label{sec:intro}

Volumetric video streaming enables users wearing AR/VR devices to view a 3D scene with 6 degrees of freedom (6-DoF) while the scene changes in time. It is expected to revolutionize both the way we communicate and entertain. Despite the great potential, the high bandwidth requirement and the computational complexity of volumetric video streaming have been major roadblocks to its wide adoption. The 3D scene needs to be transmitted and rendered in real time, which inevitably incurs high requirements in network bandwidth and computation resources.

To reduce the consumption of bandwidth and alleviate the computational burden of surface reconstruction, point clouds have been recognized as a promising 3D representation for volumetric streaming in emerging AR/VR applications~\cite{han2020vivo,lee2020groot}, given the flexibility in capturing~\cite{reimat2021cwipc} and compressing~\cite{gpccwhitepaper} point clouds. Tremendous progress has been made in the past few years in point cloud compression, including the MPEG Point Cloud Compression (PCC) standards~\cite{gpccwhitepaper} and recent deep learning based methods~\cite{wang2022sparse,zhang2023yoga,he2022density}. These methods are typically developed to compress either the geometry (point coordinates) or color, and  have demonstrated gains over the MPEG PCC standards. However, there remains a gap between the reconstruction of point cloud and the rendering of high-quality images on the clients' display. The geometry and color compression methods are mostly optimized to accurately reproduce the coordinates and colors of points, rather than the rendering quality. As a result, the rendering quality from the compressed point clouds using standard rendering tools is sometimes severely compromised with visible holes and color artifacts. Although the state-of-the-art learned point cloud rendering method~\cite{chang2023pointersect} can improve the rendering quality, it is very slow and cannot support real-time free viewpoint.

To address the limitations in existing approaches for point cloud compression and rendering, and inspired by the recent work of 3D Gaussian splatting~\cite{kerbl20233d}, we develop a novel point cloud compression scheme. Our key innovation in this paper is to design a point cloud compression scheme that compress the point cloud to a compact bit-stream that can be directly decoded to renderable 3D Gaussians, bridging the gap between point cloud compression, reconstruction, and rendering. 
Because the state-of-the-art scalable geometry coding approach in ~\cite{wang2022sparsepcgc} already achieves high coding performance for geometry,
we focus on the scalable compression of  the color as well as other information needed for  decoding the Gaussian representation.  Our method supports real-time decoding and rendering of high quality color point clouds, paving the way towards building interactive 3D streaming applications with free view points.

We achieve the goal of efficiency and high quality by introducing multiple techniques, including a hierarchical compression scheme and a novel geometry-invariant 3D sparse convolution.
The hierarchical design allows multiple levels of details at different bit-rate ranges by extracting and coding the Gaussian features following the octree structure, where the feature transformation and entropy coding at a current level are conditioned on the up-sampled features from the previous coarser level. 
Our feature extraction, compression and decompression modules are built upon the sparse 3D convolutional neural network based on the Minkowski Engine~\cite{choy20194d}. To overcome the challenge of irregularity with point cloud data, we propose a novel geometry-invariant 3D sparse convolution, which is crucial for color feature extraction and reconstruction. With the help of a fast and differentiable renderer leveraging the 3D Gaussian representation, we can end-to-end optimize the encoder and decoder jointly towards better rendering quality and lower bit-rates.

We train our models using the training set of the THuman 2.0 dataset~\cite{yu2021function4d} and evaluate them on the testing set of the THuman 2.0 dataset and the 8i Voxelized Full Bodies (8iVFB) dataset \cite{dataset8i}. Our method achieves higher perceptual quality (both in terms of PSNR and LPIPS) in the rendered images at similar bit rates as G-PCC v22 and the state-of-the-art deep learning based point cloud methods using the geometry compression model in \cite{wang2022sparsepcgc} and attribute compression method in \cite{fang20223dac} and \cite{zhang2023scalable}. 
The contributions of our work are summarized as follows:
\begin{itemize}
  \item We propose a novel end-to-end learned point cloud compression method that directly decodes bit-streams to renderable 3D Gaussians, bridging the gap between point cloud compression, reconstruction, and rendering. Leveraging a differentiable renderer utilizing Gaussian splatting, we directly optimize the rendering quality vs. bit-rate trade-off. 
  \item We adapt sparse convolution for feature extraction, squeezing, conditional entropy coding, and reconstruction. We propose a novel geometry-invariant 3D sparse convolution to address the problem of non-uniform point density in the point cloud.
  \item We propose a novel multi-resolution coding framework for compressing the color and rendering-related information. The features at a current resolution  are squeezed and entropy-coded and reconstructed conditioned on the features from the lower resolution to maximally exploit the redundancy across resolutions.
  It generates a scalable bit-stream that provides a higher level of details as the bit rate increases.
  \item Our method significantly improves the rendering quality at similar bit-rates compared to both standard and learned methods for  point cloud color compression, while substantially reducing the decoding and rendering time. 
\end{itemize}

\section{Related Works}

\subsection{Point Cloud Compression}


With the wide applications of point clouds in autonomous driving, AR/VR, and volumetric video streaming, point cloud compression has been an active research area in the past few years. The MPEG 3D Graphics Coding group has standardized the video-based (V-PCC) and the geometry-based point cloud codecs (G-PCC)~\cite{chen2023introduction,graziosi2020overview}. The V-PCC projects 3D point clouds to 2D images and encodes them by a standard video codec. It is efficient in terms of compression ratio especially for point cloud videos, but lacks level-of-detail (LoD) scalability and may introduce 3D artifacts in the reconstructed geometry. The G-PCC, on the other hand, encodes the geometry using an octree structure, with context-adaptive entropy coding. G-PCC stream is scalable, making it easy to adapt to time-varying network bandwidths in streaming applications. Given the coded geometry, it adopts the Region-Adaptive Hierarchical Transform (RAHT) to efficiently encode the color into a scalable stream.

With the advancement in neural network architectures for processing point clouds~\cite{wang2019dynamic,thomas2019kpconv,choy20194d}, neural network-based methods have shown great potential in boosting the rate-distortion performance in point cloud compression. Existing methods either focus on the compression of only point cloud geometry~\cite{huang2020octsqueeze,que2021voxelcontext,fu2022octattention,cui2023octformer,mao2022learning,wang2021lossy,wang2022sparsepcgc,he2022density} or compression of color given usually losslessly coded geometry~\cite{sheng2021deep,fang20223dac,wang2022sparse,zhang2023yoga,zhang2023scalable}. These methods demonstrate great potential in compressing point clouds to a low bit-rate while maintaining the color and geometry fidelity. However, all these methods compress geometry and color separately, unaware of the rate vs. rendering-quality trade-off, and may lead to  blurred details with standard rendering tools. In this paper, we propose a novel end-to-end learned  framework that compresses the color and other rendering information, so that the decoded bit stream along with the geometry bits can be  directly decoded  to renderable 3D Gaussians, leading to higher rate-distortion performance and efficiency.

\subsection{3D Gaussian Representation for Point Cloud Rendering}

Our work is inspired by the work in~\cite{kerbl20233d}, which demonstrates that 3D Gaussian clouds are capable of representing a 3D scene with high fidelity and rendering images without holes. Based on this, the recent work of P2ENet~\cite{hu2024low} demonstrates that point cloud can be rendered in high quality and with high speed when converted to 3D Gaussians using a pre-trained neural network without per-scene optimization. This technique has been also used in 3D scene generation tasks~\cite{xu2024agg}. Since no per-scene optimization process is needed, this technique potentially supports real-time point cloud streaming applications. However, bandwidth limit is still a major roadblock for these applications. Hence in this work, we investigate how to leverage the Gaussian splatting renderer to optimize a point cloud compression network to achieve good trade-off between rate and rendering-quality. We  develop neural network modules  to extract compact features from the point cloud, conduct entropy coding, and decode the features to 3D Gaussians for high-quality and fast rendering.

\section{Method}

\subsection{Overview of the Proposed Framework}
\begin{figure*}[t]
    \centering
    \includegraphics[width=1.0\textwidth]{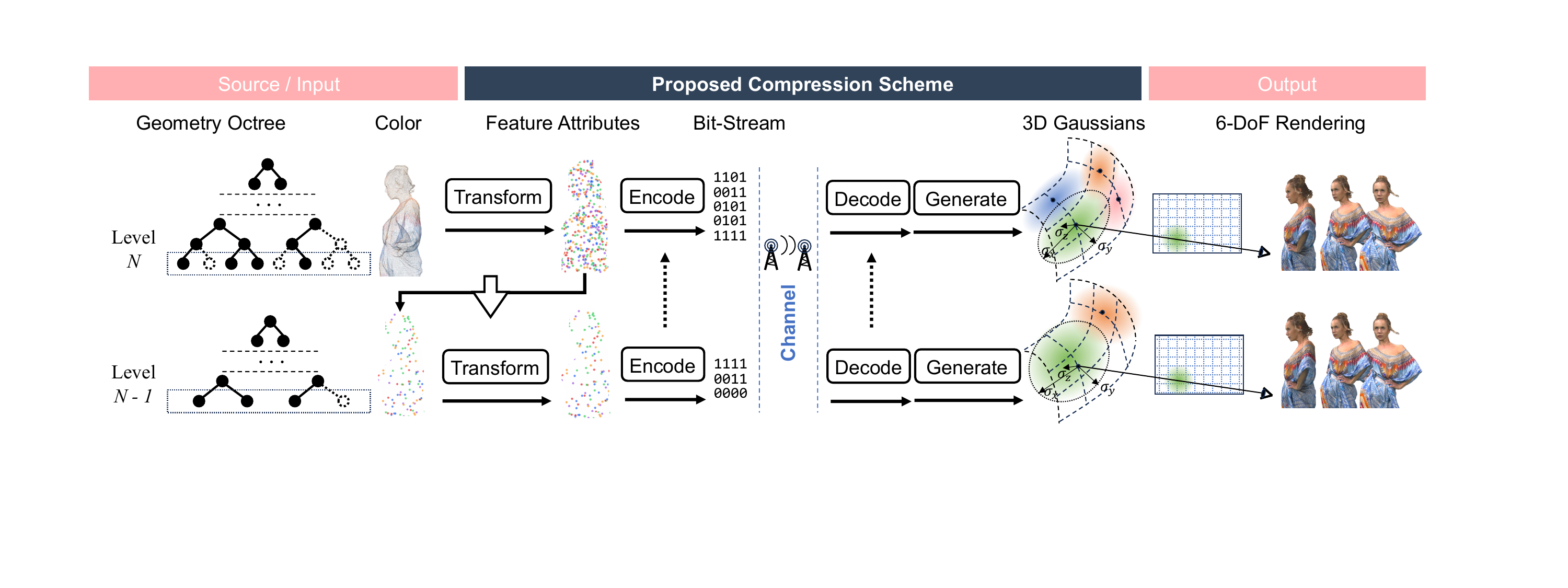}
    \caption{The proposed multi-resolution compression and rendering framework. The source point cloud is encoded into a scalable bit-stream, and delievered up to a resolution based on the sustainable network throughput. 
    The client decodes the received bit-stream to 3D Gaussians and renders the point cloud according to the client current view point. 
    }
    \label{fig:structure}
\end{figure*}



We represent the point cloud geometry (\textit{i.e.} the coordinates of all points) using the  octree structure, which has been found to be efficient in representing the sparse structure of point clouds~\cite{huang2006octree}. The octree is constructed by recursively subdividing the 3D space into 8 equal octants. Each node represents the occupancy of the corresponding octant. If one octant is completely empty, the corresponding node is marked unoccupied and not further subdivided. A point cloud where each coordinate is represented with $N$ bits can be represented by an octree with $N$ levels, where each leaf node represents an original point in the point cloud. The octree structure provides a multi-resolution representation of the geometry of the point cloud, which naturally facilitates scalable compression and rendering with multiple levels of detail. It also allows us to efficiently find the neighbors of a point in the point cloud, which is crucial for sparse 3D convolutions.

With the octree  structure, we compress the point cloud by first losslessly encode the geometry using an existing scalable point cloud geometry compression method, such as SparsePCGC~\cite{wang2022sparsepcgc} or G-PCC~\cite{gpccwhitepaper}, and then encode the color and other rendering related information given the geometry defined by the octree structure. The proposed compression and rendering pipeline is illustrated in Fig.~\ref{fig:structure}.


We first extract color features \footnote{For simplicity, we call them color features. But they actually carry additional information to recover the 3D Gaussians for rendering.}  at all leaf nodes at level $N$ of the octree representation of a point cloud.  We then generate the hierarchical feature representation by repeatedly downsampling the extracted features following the octree structure, until a chosen level $L$, using the average pooling in a $2\times 2 \times 2$ neighborhood. We then encode the features at different levels using a hierarchical  coding scheme.
We start with coding the color features at level $L$ independently, and then code features at  successive higher levels conditioned on the upsampled features from the lower level, until level $N$, using the conditional feature transform and entropy coding modules shown in Fig.~\ref{fig:structure}. 

The hierarchical  coding scheme naturally generates a scalable bit-stream, and  allows us to achieve different bit-rates and different levels of details, by transmitting up to different levels $M$  ($L \leq M \leq N$),  with $M$ chosen based on the sustainable network throughput between the sender and the receiver or the decoder processing constraint. 
At the receiver, the decoded features at level $M$ are then converted to Gaussian parameters using the Gaussian Generation module in Fig.~\ref{fig:structure}.

The individual modules
are explained in more detail  below. The network architectures used for different modules are given in the supplementary material.

\subsection{Full-Resolution feature Extraction}
We extract features at level $N$ using a multi-layer Minkowski Convolution Network~\cite{choy20194d}. The original Minkowski convolution is defined as,
\begin{equation}
  \mathbf{x}_u = \sum_{u+i \in \mathcal{N}(u, K, P^{\text{in}})} W_i \mathbf{x}_{u+i} \text{ for } u \in P^{\text{out}}, 
\end{equation}
where  $\mathcal{N}(u, K, P^{\text{in}})$ is the set of occupied input points that are within the  $K\times K\times K$ neighborhood of the output point $u$. $P^{\text{in}}$ and $P^{\text{out}}$ denote the set of occupied points.  We observe that this definition of convolution is not geometry invariant, as it tends to produce higher magnitude where the neighborhood around $u$ is denser. To address this problem, we propose a {\it geometry-invariant 3D sparse convolution}, which is crucial for  color feature extraction and reconstruction. The geometry-invariant 3D sparse convolution is defined as,
\begin{equation}
\begin{split}
& \mathbf{x}_u = \left( \sum_{u+i \in \mathcal{N}(u, K, C^{\text{in}})}   W_i^2 \right)^{-\frac{1}{2}} \sum_{u+i \in \mathcal{N}(u, K, C^{\text{in}})} W_i \mathbf{x}_{u+i} \\ & \text{ for } u \in P^{out},
\end{split}
\end{equation}
which normalizes the output response by the $L_2$ norm of {\it active} kernel weights. This normalization ensures that the convolution response is consistent across different point densities in the local area. 

\subsection{Conditional Transform and Entropy Coding}
\label{sec:entropy_coding}
\begin{figure*}[t]
    \centering
    \includegraphics[width=0.9\textwidth]{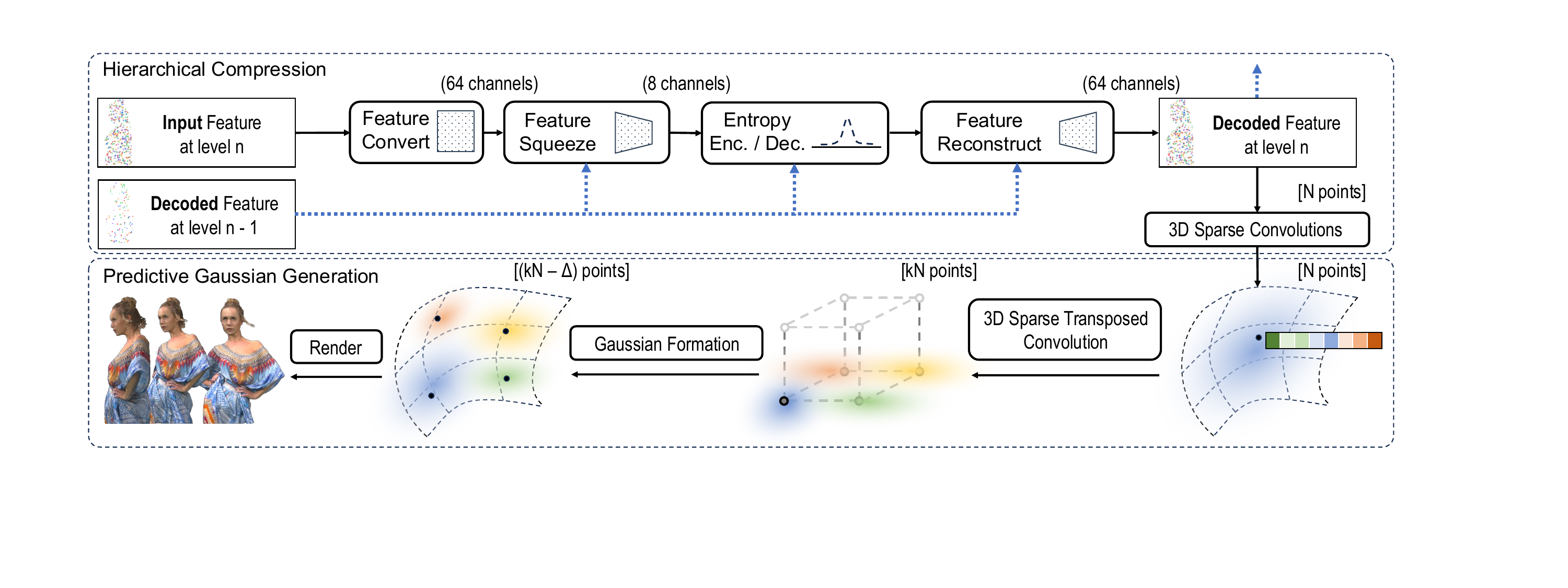}
    \caption{Illustration of key components of the proposed method.
    }
    \label{fig:entropy_coding}
    \vspace{-4mm}
\end{figure*}


We compress the 
features at each level using feature squeezing and conditional entropy coding, shown in Fig.~\ref{fig:entropy_coding}. 
Suppose we are coding  {input} features at level $n$, given {decoded} features at level $n - 1$. 
%
%
We first apply a feature squeezing module to reduce the number of output feature channels from 64 to 8, by processing both the 64 input channels at level $n$ and the 64 context channels obtained by upsampling the features from level $n-1$. 
The motivation of feature squeezing is that the features at  level $n$ can usually be predicted well from features at level $n-1$ and we only need to encode the \textit{residual} information in level $n$, which is usually sparser and requires fewer channels to represent. The squeezing module learns to generate the \textit{residual} in a general sense given the context, and prepare the residual features for quantization and entropy coding.
We use the Minkowski Inception ResNet architecture~\cite{wang2021lossy} with the proposed geometry-invariant 3D convolution for the squeezing module.

To encode the squeezed features, we first quantize the feature components to integers, and use a conditional arithmetic coder to encode the quantized features to binary bit-streams using the up-sampled features from  level $n - 1$ as the conditioning context. During training, the quantization is simulated using additive uniform noise $U(-0.5, 0.5)$. Following the entropy coding method of \cite{balle2018variational}, we assume the all feature values are independent and each follow a Gaussian distribution with spatially varying mean and variance. Let $x_u^d$ represent the $d$-th component of the $D$ dimensional feature $x_u$ at point $u$. The probability of $x_u^d$   can be calculated as
\begin{equation}
  p(x_u^d) = \Phi(x_u^d + 0.5; \mu_u^d, \sigma_u^d) - \Phi(x_u^d - 0.5; \mu_u^d, \sigma_u^d),
\end{equation}
where $\Phi$ is the Gaussian cumulative distribution function (CDF) and $\mu_u^d, \sigma_u^d$ are estimated from the context features. We assume the number of bits needed to code $x_u^d$ equals the negative log likelihood, i.e.,
\begin{equation}
  R(x_u^d) = -\log_2 p(x_u^d).
  \label{eq:code-length}
\end{equation}
The bit-rate loss at level $n$ is defined as the sum of the bits for all points at this level $P_n$:
\begin{equation}
  \mathcal{L}_{R}^n = - \sum_{u \in P_n} \sum_{d=1}^D  \log_2 p(x_u^d).
  \label{eq:rate_loss}
\end{equation}
During inference, the  arithmetic coder will utilize the estimated probability $p(x_u^d)$ from the context features to encode $x_u^d$, with a code length very close to (\ref{eq:code-length}) . 


After entropy decoding, the quantized features are concatenated with the context features along the channel dimension, and go through the feature reconstruction module. This module produces the {decoded} features at level $n$, which serve two purposes: (1) They can be used as the context to encode the next level $n + 1$; (2) If we need to render at level $n$, we can generate the 3D Gaussian cloud directly from the reconstructed features. By encoding and decoding recursively across the scale using the proposed scheme, we  achieve level-of-detail scalability, while reducing the total bit rate up to level $N$.

\subsection{Predictive 3D Gaussian Generation}
\label{sec:predictive_rendering}


The reconstructed features we obtain at each level are feature vectors attached to occupied grid points in the voxelized space. 
We convert the feature vectors at the highest level $M$ that is transmitted and received to renderable 3D Gaussians.  Each 3D Gaussian is parameterized by a center coordinate (\textit{a.k.a.} the 3D means) $(\mu_x, \mu_y, \mu_z)$, a covariance matrix $\Sigma$, an opacity value $o$, and a color triplet $c \in [0, 1]^3$. The covariance matrix is parameterized as $\Sigma = \mathbf{R}^T \mathbf{S}^T \mathbf{S} \mathbf{R}$, where $\mathbf{R}$ is a rotation matrix and $\mathbf{S} = \text{diag}(\sigma_x, \sigma_y, \sigma_z)$ is a diagonal matrix. The rotation matrix $\mathbf{R}$ is calculated from a quaternion $\mathbf{q} = (q_w, q_x, q_y, q_z)$.  Since the features are extracted from the original point cloud that does not have anisotropic color, we simplify the original parameterization in \cite{kerbl20233d} by using a constant color $c \in [0, 1]^3$  for each Gaussian,  instead of multiple coefficients of the spherical harmonics. The conversion from the reconstructed features to 3D Gaussians is in effect generating the following 14 parameters, $\mu_x, \mu_y, \mu_z, \sigma_x, \sigma_y, \sigma_z, q_w, q_x, q_y, q_z, o, c$, for all non-empty points. Each 3D Gaussian is rendered to the screen space using the  differentiable Gaussian splatting renderer proposed in \cite{kerbl20233d}, and the pixels covered by overlapping Gaussian splats are given a color using alpha blending following \cite{hu2024low}.

A simple approach for this conversion is just to use the grid coordinates of  the points  as the 3D means, and predict the remaining parameters from the features. However, this simple approach has two limitations. First, the number of grid points decrease exponentially with the level of the octree. At a coarse level there might not be enough Gaussians to render the fine-grained texture of the 3D scene. Second, at a coarse level the grid coordinates may not be the correct center positions due to octree-based downsampling. Therefore, we propose a Gaussian Generation module, shown in Fig.~\ref{fig:entropy_coding}. The key idea is to generate more points from the original grid points, while allowing the module learning to disable falsely generated points by setting the opacity $o \to 0$. For each grid point with a feature vector, we generate features at $k=8$ sub-points using the generative transpose 3D convolution shown in Fig.~\ref{fig:entropy_coding}. The features at all sub-points are then converted to  predicted Gaussian parameters, including the coordinate offsets $(\delta_x, \delta_y, \delta_z)$, and other parameters $\sigma_x, \sigma_y, \sigma_z, q_w, q_x, q_y, q_z, o, c$. The coordinate offsets are added to the grid point coordinates to produce the 3D means $(\mu_x, \mu_y, \mu_z)$. Using the information carried by the feature, the module learns to enable or disable (by setting opacity value $o$ close to 0) the sub-points and move them to the correct positions through the predicted offsets to represent the underlying textured smooth surface. In Fig.~\ref{fig:entropy_coding}, $\delta$ indicates the number of Gaussians with $o$ close to 0.

In practice for $N=10$, we found that this generative prediction is helpful at level $M=N - 2$ or lower. But at level $N - 1$ and $N$, we just need to predict one Gaussian  (i.e. $k=1$) for each occupied point while setting the opacity of all Gaussians to a constant of 1. For each Gaussian, we  predict  the offset from the original  grid coordinates  and other parameters including $\sigma_x, \sigma_y, \sigma_z, q_w, q_x, q_y, q_z, c$.

\subsection{Training}
\label{sec:training}

With the hierarchical coding structure and the scalable delivery and rendering scheme, we encode the point cloud into bit-streams at levels $L$ to $N$, and depending on the target bit rate, deliver from level $L$ up to level $M$ and render at level $M$, with $L\leq M \leq N$. 

Because we want to design a scalable coding scheme that can accommodate a large rate range corresponding to deliver up to  a level between $M_{\min}$ and $M_{\max}$, 
we train the model using the rate-distortion loss function defined on all coding and rendering levels, as,
\begin{equation}
  \mathcal{L} = \sum_{n\in \{L, \cdots, M_{\max} \}} \mathcal{L}_{R}^n + \lambda \sum_{n \in \{M_{\min}, \cdots,  M_{\max}\}} \mathcal{L}_{D}^n,
\end{equation}
where the bit-rate term $\mathcal{L}_{R}^n$ is defined as in Eq.~(\ref{eq:rate_loss}) for level $n$. The distortion loss $\mathcal{L}_{D}^n$ includes the weighted L1 distance, the SSIM loss, and the LPIPS loss~\cite{zhang2018unreasonable}, calculated between the rendered images $\hat{x}^n$ at level $n$ and the ground truth views $x$ obtained by rasterizing the mesh provided in the training dataset, defined as
\begin{equation}
  \mathcal{L}_{D}^n = \alpha ||x - \hat{x}^n||_1 + \beta \left(1 - \text{SSIM}(x,\hat{x}^n) \right) + \gamma \mathcal{L}_{\text{LPIPS}}(x, \hat{x}^n).  
\end{equation}
During training, we calculate the rendering distortion averaged from  4 random view points for each point cloud in the training set.
We set $\alpha = 3$, $\beta = 0.2$, and $\gamma = 1$ in our experiments. Note that to achieve more fine-grained rate control, besides the level scalability, we also train different models with different $\lambda$. We set $\lambda \in \{5, 10\}$ in our experiments. It thus produces 2 sets of scalable bit-streams.

We have observed that, when the level of details is too sparse (\textit{i.e}, $M$ is too small), the features do not carry sufficient information for rendering with adequate quality. We found that for a typical 10-bit point cloud ($N = 10$), the best rate-quality trade-off is achieved by transmitting starting from level $L=7$ up to level $M$ with $M_{\min}=8, M_{\max}=9$. Although it is possible to transmit all way to level $10$, the quality improvement from level $9$ to level $10$ does not justify the additional bits needed.
Thus each scalable bit stream obtained by a model trained with a given $\lambda$  has  2 rate-quality points (corresponding to $M=8, 9$), realizing a total of 4 Rate-Quality points. When decoding at $M=8$, we use transpose convolution to generate 8 Gaussians for each point. But at $M=9$, we only generate one Gaussian per point. We use the Adam optimizer with a learning rate of $10^{-4}$ and a batch size of 4. We train the model for 60,000 iterations on a single NVIDIA A100 GPU. We will release our code and the trained model to the public upon acceptance.

\section{Experiments}

\begin{figure*}[t]
    \centering
    \begin{subfigure}{0.24\textwidth}
      \includegraphics[width=\textwidth]{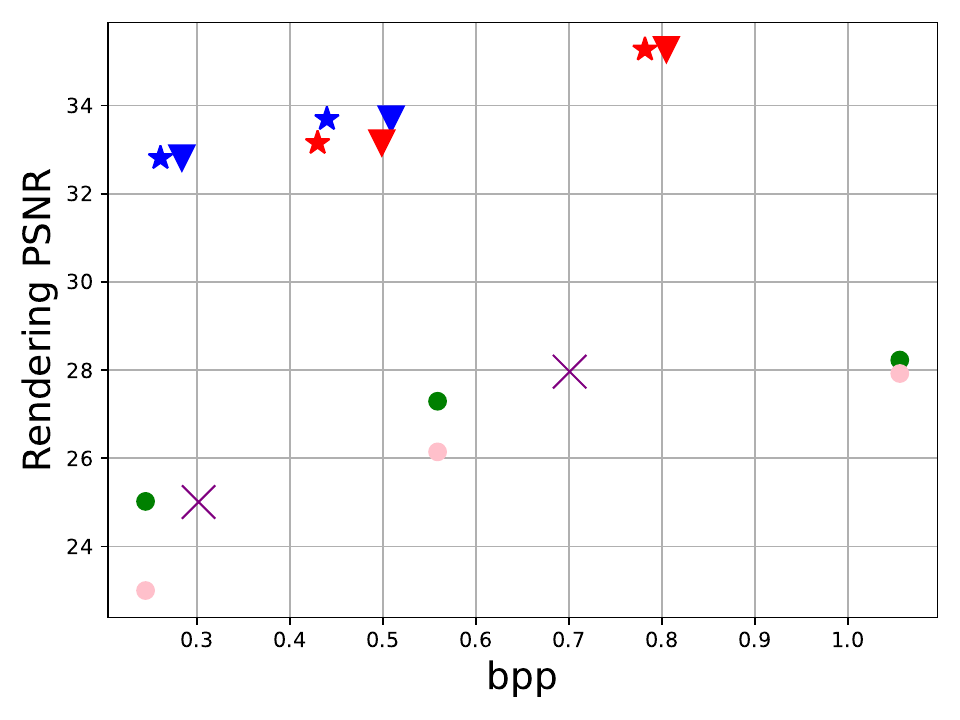}
      \caption{THuman 2.0, PSNR$\uparrow$.}
      \label{fig:thuman_psnr}
    \end{subfigure}
    \begin{subfigure}{0.24\textwidth}
      \includegraphics[width=\textwidth]{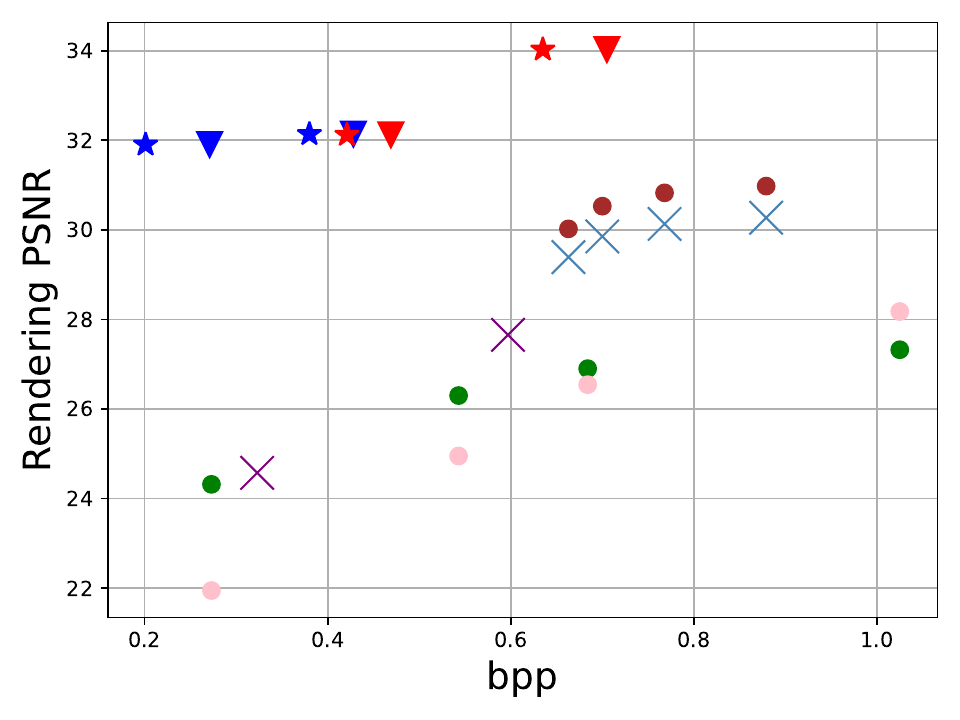}
      \caption{8iVFB, PSNR$\uparrow$.}
      \label{fig:8i_psnr}
    \end{subfigure}
    \begin{subfigure}{0.24\textwidth}
      \includegraphics[width=\textwidth]{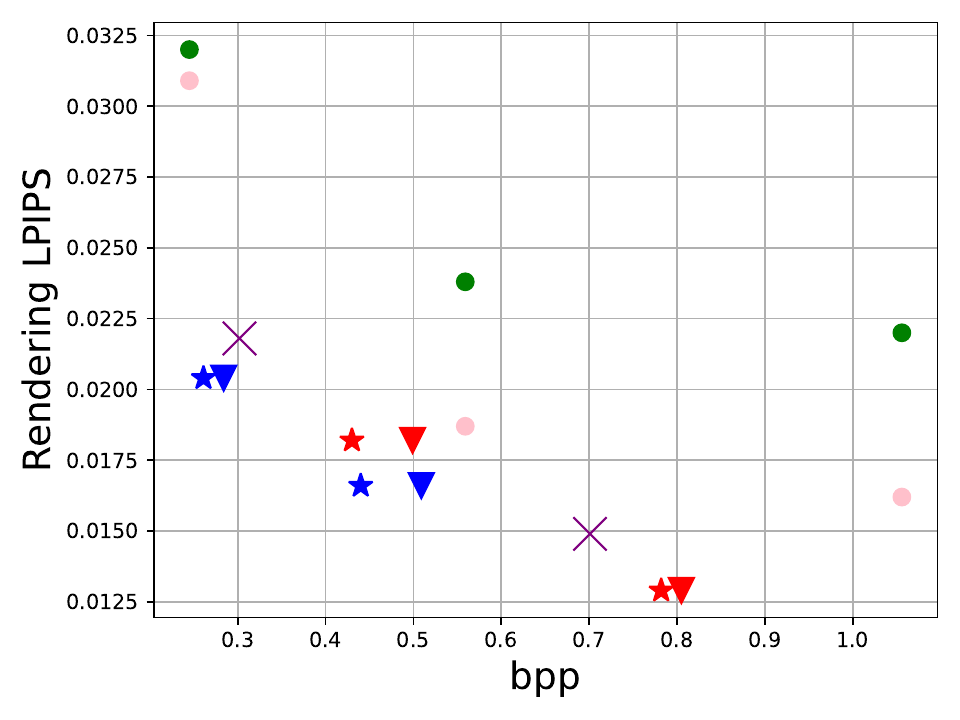}
      \caption{THuman 2.0, LPIPS$\downarrow$.}
      \label{fig:thuman_lpips}
    \end{subfigure}
    \begin{subfigure}{0.24\textwidth}
      \includegraphics[width=\textwidth]{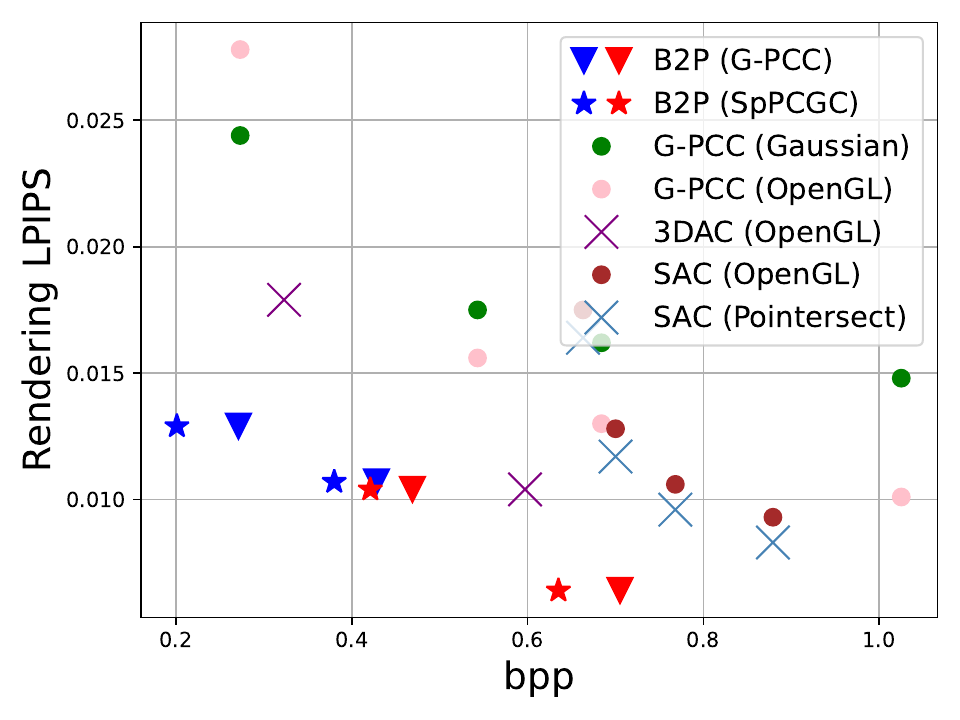}
      \caption{8iVFB, LPIPS$\downarrow$.}
      \label{fig:8i_lpips}
    \end{subfigure}
    \caption{Rendering distortion at different bit-rate by the proposed Bits-to-Photon (B2P) and baseline methods. We train 2 models with 2 different $\lambda$ values. Each model has 2 levels of scalability, forming 2 groups of scalable R-D points, plotted using different colors. Other methods are not scalable.}
    \label{fig:rate-distortion}
  \end{figure*}

\subsection{Settings}

We leverage the THuman 2.0 dataset~\cite{yu2021function4d} for training and testing, and also evaluate our trained models on the 8i Voxelized Full Bodies (8iVFB) dataset \cite{dataset8i}. The THuman 2.0 dataset provides 525 meshes built from RGBD captures with human subjects performing actions. We densely sample the meshes to obtain the color point clouds, and use the point clouds to train and evaluate our model. The dataset is divided into a training set of 500 meshes $(0000 - 0499$) and the rest for testing, this guarantees a cross-subject validation with the testing set containing only unseen subjects. In our experiments, we report results with 8 point clouds from unrepeating subjects, namely $\{0507, 0509, 0511, 0513, 0515, 0517, 0519, 0521\}$. To evaluate the generalizability, we also test on the MPEG CTC 8iVFB dataset, which provides 4 full-body human point clouds in bit-depth 10, which can be represented losslessly by a 10-level octree.

We evaluate the following compression schemes:
\begin{itemize}
  \item \textbf{G-PCC}. We use the MPEG G-PCC reference software TMC13v22~\cite{tmc13} to encode both the geometry (using octree) and the color (using RAHT). It varies the geometry quantization scale and color quantization parameters to achieve different bit-rates. Therefore the resulting bit points are not part of a scalable stream. We use the default settings for geometry and color quantization.
  \item \textbf{3DAC}. We use the state-of-the-art learned scalable point cloud geometry compression method SparsePCGC~\cite{wang2022sparsepcgc} to encode the geometry, and use the learned point cloud attribute compression method 3DAC~\cite{fang20223dac} to encode the color.
  \item \textbf{SAC}. Similarly we use SparsePCGC to encode the geometry and use the learned scalable point cloud attribute compression (SAC) model~\cite{zhang2023scalable} to encode the color.\footnote{Since the code is not publicly available, we use the bit-stream and reconstructions kindly shared by the authors, which only include point clouds in the 8iVFB dataset with lossless geometry.}
  \item \textbf{B2P (G-PCC)}. We use our method to compress the color and other rendering information, given geometry coded by G-PCC up to the same octree level as the color bit-stream.  required level. We use the same settings as the G-PCC method for geometry coding.
  \item \textbf{B2P (SpPCGC)}. Same as above but we replace the geometry coding with SparsePCGC~\cite{wang2022sparsepcgc}.
\end{itemize}

\begin{figure}[t]
    \begin{subfigure}{0.24\linewidth}
        \includegraphics[width=\linewidth]{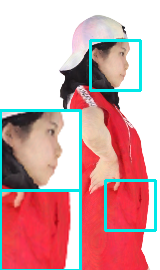}
        \centering
    \end{subfigure}
    \begin{subfigure}{0.24\linewidth}
        \includegraphics[width=\linewidth]{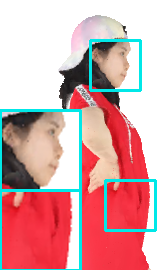}
    \centering
    \end{subfigure}
    \begin{subfigure}{0.24\linewidth}
        \includegraphics[width=\linewidth]{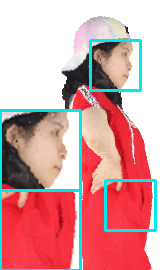}
    \centering
    \end{subfigure}
    \begin{subfigure}{0.24\linewidth}
        \includegraphics[width=\linewidth]{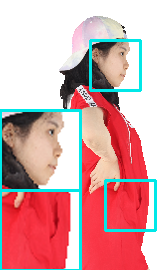}
        \centering
    \end{subfigure}

    \centering
    \begin{subfigure}{0.24\linewidth}
    \centering
        \scriptsize B2P \\ (0.93 bpp)
    \end{subfigure}
    \begin{subfigure}{0.24\linewidth}
    \centering
        \scriptsize G-PCC (1.00 bpp)\\(OpenGL)
    \end{subfigure}
    \begin{subfigure}{0.24\linewidth}
        \centering
        \scriptsize 3DAC (1.18 bpp)\\(OpenGL)
    \end{subfigure}
    \begin{subfigure}{0.24\linewidth}
        \centering
        \scriptsize Original \\(GT Mesh)
    \end{subfigure}
    \caption{Visual results on the decoded and rendered point cloud on the THuman 2.0 dataset, compared to images rendered from the ground truth mesh.}
    \label{fig:visual_thuman}
    \vspace{-5mm}
\end{figure}

We report bit-rates in bit-per-point (bpp), using the number of points from the original point cloud. We evaluate the rendering distortion at different bit-rates for comparison. We average the rendering distortion over 12 camera views forming a circle with equal angle intervals around the subject. Our decoder operation includes entropy decoding the squeezed features,  reconstructing the features, and generating the Gaussian parameters, which can then be used to efficiently produce the  rendered images  given the camera extrinsics and intrinsics. For the G-PCC and the learned baseline, we first decode the point cloud, and then render with the existing standard point cloud renderer provided by OpenGL and packaged in Open3D~\cite{zhou2018open3d}, which rasterizes each point as a solid square in the camera plane. We set the square size according to the average point density so there is no visible hole. We also include an alternative renderer, taking each point as a spherical Gaussian with a fixed diagonal covariance matrix $\Sigma = \text{diag}(\sigma, \sigma, \sigma)$, where $\sigma = \bar{d}$ corresponding to the average nearest point distance. Since THuman 2.0 dataset provides ground truth meshes, we evaluate the rendering quality using the PSNR, MS-SSIM, and the LPIPS~\cite{zhang2018unreasonable} between the rendered images from the decoded point clouds and the same views rendered from the ground truth meshes using standard rasterization. For the 8iVFB dataset which does not provide ground truth meshes, we first use the Screened Poisson surface reconstruction~\cite{kazhdan2013screened} to convert the point clouds to the meshes, and then render the meshes to produce ground truth images.

\subsection{Rate-Distortion Performance}

\begin{figure}[t]
    \centering


    \begin{subfigure}{0.24\linewidth}
        \includegraphics[width=\linewidth]{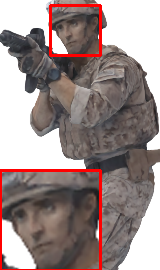}
        \centering
    \end{subfigure}
    \begin{subfigure}{0.24\linewidth}
        \includegraphics[width=\linewidth]{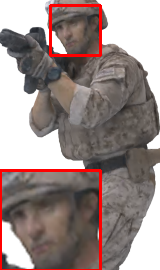}
    \centering
    \end{subfigure}
    \begin{subfigure}{0.24\linewidth}
        \includegraphics[width=\linewidth]{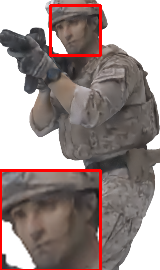}
        \centering
    \end{subfigure}
    \begin{subfigure}{0.24\linewidth}
        \includegraphics[width=\linewidth]{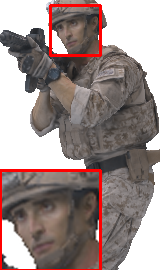}
        \centering
    \end{subfigure}

    \centering
    \begin{subfigure}{0.24\linewidth}
    \centering
        \scriptsize B2P\\ (0.59 bpp)
    \end{subfigure}
    \begin{subfigure}{0.24\linewidth}
    \centering
        \scriptsize G-PCC (0.62 bpp)\\(OpenGL)
    \end{subfigure}
    \begin{subfigure}{0.24\linewidth}
        \centering
        \scriptsize SAC (0.64 bpp)\\(OpenGL)
    \end{subfigure}
    \begin{subfigure}{0.24\linewidth}
        \centering
        \scriptsize Original \\(Poisson Mesh)
    \end{subfigure}
    \caption{Visual results on the decoded and rendered point clouds on the 8iVFB dataset, compared to images rendered using the generated meshes from original point clouds.}
    \label{fig:visual_8i}
    \vspace{-5mm}
\end{figure}

\begin{table*}[t]
    \centering
    \scriptsize
    \caption{Bit-rate and rendering quality at different octree levels and with different methods. We show averaged results over the 4 point clouds in the 8iVFB dataset.}
    \begin{tabular}{c|c|c|c|c|c|c|c|c|c}
      \toprule
      \multirow{2}{*}{Codec} & \multirow{2}{*}{Level} & \multirow{2}{*}{Renderer} & \multirow{2}{*}{bpp}  & \multirow{2}{*}{PSNR $\uparrow$} & \multirow{2}{*}{LPIPS $\downarrow$} &  \multirow{2}{*}{MS-SSIM $\uparrow$} & \multicolumn{2}{c|}{Decode Time} & Render \strut \\
      & & & & & & &  Geometry & Color & Time \\
      \midrule
      B2P (SpPCGC) & 8 & -        & 0.45 & \textbf{32.1} & \textbf{0.010} & \textbf{0.9948} & 270 ms & 26 ms & 4 ms \strut \\
      G-PCC & 9 & Global Gaussian & 0.54 & 25.0 & 0.016 & 0.9824 &  \multicolumn{2}{c|}{2 sec} & 3 ms \strut\\
      G-PCC & 9 & Pointersect & 0.54 & 29.4 & \textbf{0.010} & 0.9934 &\multicolumn{2}{c|}{2 sec} & 1 sec \strut\\
      
      3DAC & 9 & OpenGL & 0.60 & 27.7 & \textbf{0.010} & 0.9893 & 370 ms & 1 sec & 3 ms \strut \\
      \midrule
      B2P (SpPCGC) & 9 & -        & 0.66 & \textbf{34.0} & \textbf{0.006} & \textbf{0.9967} & 370 ms & 73 ms & 5 ms \strut \\
      G-PCC & 9+\footnotemark[3] & Global Gaussian & 0.68 & 26.5 & 0.016  & 0.9854 &  \multicolumn{2}{c|}{4 sec} & 3 ms \strut\\
      G-PCC & 9+\footnotemark[3] & Pointersect & 0.68 & 29.8 & 0.010 & 0.9939 & \multicolumn{2}{c|}{4 sec} & 1 sec \strut \\
      \bottomrule
    \end{tabular}
    \label{tab:tradeoff}
    \end{table*}

We plot the R-D points achieved by the proposed Bits-to-Photon (B2P) and compared methods in Fig.~\ref{fig:rate-distortion}. As shown by the PSNR evaluation, images rendered by our method from our decoded bit-stream have more pixel-level fidelity to the ground truth images rendered from the mesh, with over 4 dB improvements in PSNR at the same level of bit-rate over the compared methods. We also show the results using LPIPS  \cite{zhang2018unreasonable} as the distortion metric, which is more sensitive to texture similarity and more tolerant to pixel-wise difference. Despite the disadvantage of Gaussian based renderer over OpenGL in terms of sharpness, our method still achieves better rate-distortion tradeoffs.
In Fig.~\ref{fig:visual_thuman} and \ref{fig:visual_8i}, we demonstrate better visual quality with finer facial details, sharper edges and higher contrast at a lower bit-rate, compared to G-PCC, 3DAC, an SAC.


\subsection{Bit-rate, Complexity, and Distortion Tradeoff}

\footnotetext[3]{Using geometry quantization scale $0.75$.}

Table~\ref{tab:tradeoff} compares  rate, rendering quality and complexity (in terms of running time) of different methods with the 8iVFB dataset. Our evaluation is conducted on a desktop computer with an Intel i7-9700K CPU and an NVIDIA RTX 4080 Super GPU. B2P, SparsePCGC, Global-Parameter Gaussian-based rendering, 3DAC and Pointersect are using both CPU and GPU, where the neural networks parts are running on GPU. G-PCC is running on CPU.
%
As shown, B2P shows better R-D performance than compared methods at a lower decoding latency. We have the following observations:
\begin{itemize}
    \item B2P decoding and rendering at octree level 8 already achieves better R-D performance than G-PCC at level 9. Although given a better point cloud renderer (Pointersect), the rendering distortion can be further reduced with the same reconstructed point cloud, the rendering complexity is too high for  real-time 6-DoF viewing.
    \item Learning based codec achieves better R-D performance than G-PCC, but the color coding part (3DAC) is still very complex.
    \item B2P provides better rendering quality by further decode one more level in the octree, as an enhancement layer. 
\end{itemize}


\subsection{Ablation Study}

\begin{figure}[t]
    \centering
    \begin{subfigure}{0.45\linewidth}
      \includegraphics[width=\linewidth]{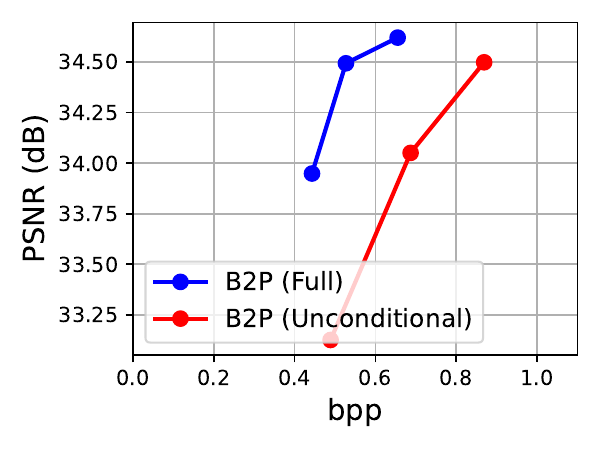}
      \caption{8iVFB, PSNR$\uparrow$.}
      \label{fig:ablation_psnr}
    \end{subfigure}
    \begin{subfigure}{0.45\linewidth}
      \includegraphics[width=\linewidth]{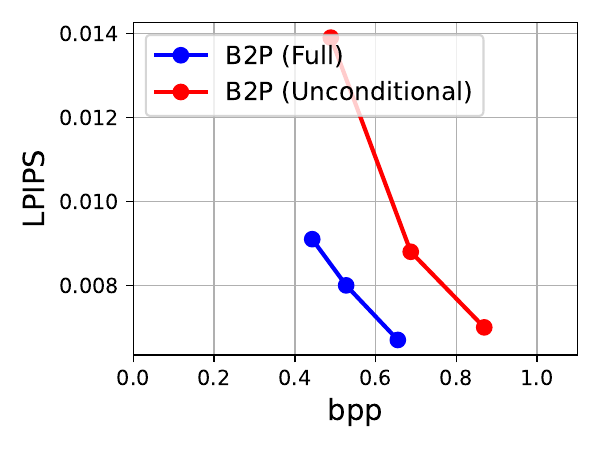}
      \caption{8iVFB, LPIPS$\uparrow$.}
      \label{fig:ablation_lpips}
    \end{subfigure}
    \caption{Ablation study on the conditional feature squeeze and reconstruction modules. With B2P (Unconditional), we remove the conditioning in both the feature squeezing and reconstruction module from the coarser layer.}
    \label{fig:ablation}
      \vspace{-4mm}
  \end{figure}

 Unlike existing work of learned point cloud attribute compression that either rely on G-PCC to compress a base layer point cloud attributes~\cite{zhang2023scalable,zhang2023yoga} or only has conditional entropy coding~\cite{wang2022sparse}, we adopt a hierarchical compression scheme where not only the entropy coding module but also the feature squeezing and reconstruction modules are conditioned on a coarser layer that has been already coded. We conduct an ablation study to show the efficacy of this technique by comparing the full scheme to a version disabling the conditional feature squeezing and reconstruction. As shown in Fig.~\ref{fig:ablation}, the conditional feature squeezing and reconstruction modules greatly contributed to the rate-distortion performance by reducing more than 25\% of the bit-rate at the same level of reconstruction quality. This technique can be also adopted by other learned scalable point cloud compression schemes for potential performance boost.





\section{Conclusion}
To address the critical roadblock in using point cloud for real-time volumetric video streaming, we propose a novel point cloud compression scheme, Bits-to-Photon (B2P), which jointly designs a scalable color compression scheme and a decoder to directly generate 3D Gaussian parameters for rendering. Leveraging  a differentiable Gaussian splatting renderer, we perform end-to-end optimization considering both compression ratio and rendering quality. We show that B2P achieves better bit-rate vs. rendering-quality performance than the state-of-the-art methods, while  substantially reducing decoding and rendering latency. We believe that B2P opens an avenue of point cloud compression optimized for rendering, and is a promising step towards developing real-time 6-DoF volumetric streaming systems.

We have only demonstrated  promising results with 2 scalable levels so far even though the framework is capable of supporting more levels.  It may be desirable to achieve a more fine-grained scalability between level 9 and 10. This can be potentially achieved by a region-adaptive coding scheme, which allocates more bits and produces more Gaussians in areas where more points are present at level 10. 
Furthermore, we have only considered coding a single point cloud. Future research will consider coding  point cloud video by exploiting decoded features in previous frames as temporal context for feature squeezing, entropy coding and reconstruction.

\section*{Acknowledgement}
This material is based upon work supported by the National Science Foundation under Grant No. 2312839.

\bibliography{aaai25}

\clearpage

\section{Appendix}
\subsection{Network Architecture}
\label{sec:intro}
We provide details on the sparse 3D convolution-based network architecture used in our method. This includes the \textit{Feature Convert} module, the \textit{Feature Squeeze} module, the \textit{Predictive Entropy Model}, the \textit{Feature Reconstruct} module, and the \textit{3D Gaussian Generation} module. Please refer to Fig.~2 (\textit{Feature Convert}, \textit{Feature Squeeze} and \textit{Predictive Entropy Model}) and Fig.~3 (\textit{3D Gaussian Generation}) in the main paper for how these modules are incorporated in the proposed compression scheme.

\subsubsection{Buidling Blocks}

\begin{figure*}[t]
  \centering
  \begin{subfigure}{0.58\linewidth}
      \centering
      \includegraphics[width=0.9\textwidth]{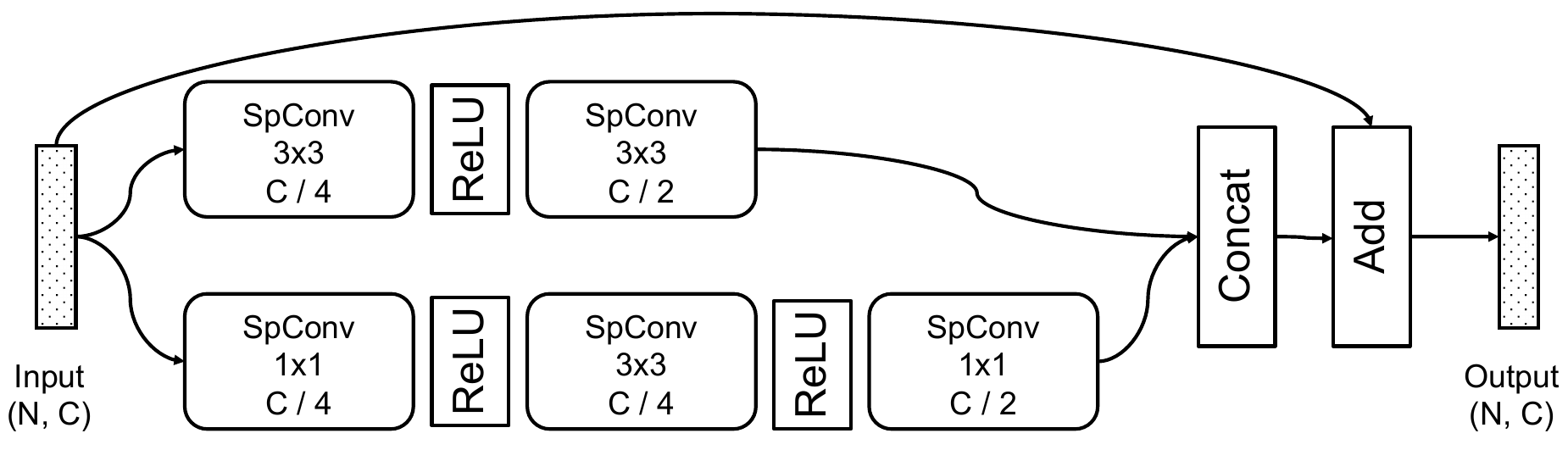}
      \caption{InceptionResBlock}
  \end{subfigure}
  \begin{subfigure}{0.38\linewidth}
      \centering
      \includegraphics[width=0.9\textwidth]{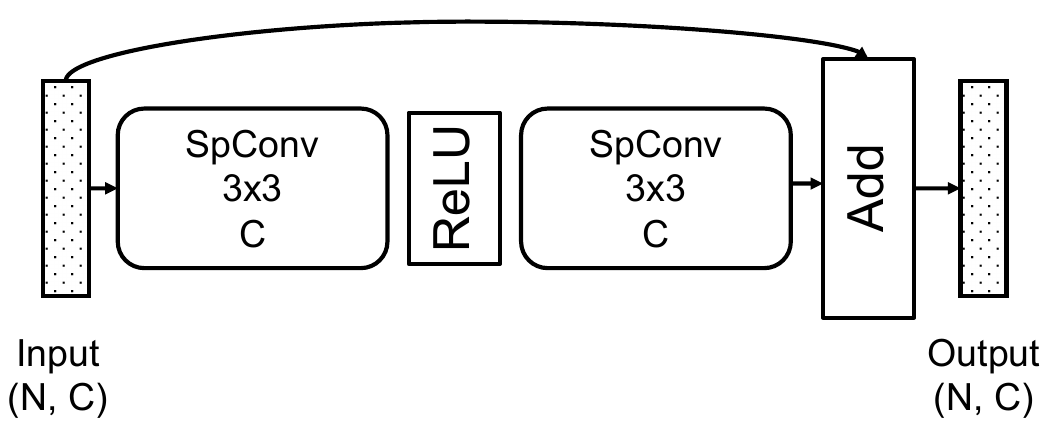}
      \caption{ResBlock}
  \end{subfigure}
  \caption{The layer-wise details of the InceptionResBlock and ResBlock. \textit{SpConv} is the abbreviation for 3D sparse convolution (Minkowski Convolution~\cite{choy20194d}). $C$ is the number of input channels and $N$ is the number of points in the input. The batch dimension is omitted for simplicity.}
  \label{fig:inception_resnet_resblock}
\end{figure*}

The proposed architecture includes two kinds of building blocks, \textit{i.e.} InceptionResBlock and ResBlock. We adopt the InceptionResBlock architecture from the work~\cite{wang2021lossy}. For completeness, we also provide the detail in this section. The ResBlock is a 3D sparse version of the original residual block~\cite{he2016deep}. The layer-wise details of these two blocks are shown in Fig.~\ref{fig:inception_resnet_resblock}.

\subsubsection{Feature Convert Module}


The \textit{Feature Convert} module architecture shown in Table~\ref{tab:feature_convert} is used for both extracting feature from the input point cloud (RGB colors as features) or converting a down-sampled point cloud with features from a denser level to features for this level. The output will 1) go to the \textit{Feature Squeeze} module for further processing, and 2) be down-sampled to the next level for further processing by another \textit{Feature Squeeze} module.

\subsubsection{Feature Squeeze Module}


The \textit{Feature Squeeze} module architecture shown in Table~\ref{tab:feature_squeeze} is used for reducing the dimensionality of the features at this level. Since we want to remove the predictable information to save bit-rates, we design the process to be conditioned on the up-sampled feature from a coarser level. Hence, the input layer in Table~\ref{tab:feature_squeeze} has $64 + 64$ input channels, corresponding to the features at this layer and the context from a coarser layer, respectively. The output will be quantized and entropy coded, with a \textit{Predictive Entropy Model}.

\subsubsection{Predictive Entropy Model}

The \textit{Predictive Entropy Model} architecture shown in Table~\ref{tab:entropy_model} is used for modeling the conditional entropy of the quantized features. We assume the squeezed features follow a conditional Gaussian distribution. The \textit{Feature Squeeze} module thus takes the upsampled feature from a coarser level as input, and generates the mean and scale of the Gaussian distribution for each point, which is used to form the cumulative distribution function (CDF) for entropy coding. In Table~\ref{tab:entropy_model}, Layer 4 generates the mean $\mu$ while Layer 5 generates the scale $\sigma$. The output will be used for entropy coding the quantized features from the \textit{Feature Squeeze} module.

\subsubsection{Feature Reconstruct Module}

The \textit{Feature Reconstruct} module architecture shown in Table~\ref{tab:feature_reconstruct} is for reconstructing the higher dimensional feature from the squeezed and quantized features given by the \textit{Feature Squeeze} module. Since the squeezing removes the predictable information, in the reconstruction the module should take the up-sampled feature from a coarser level as a conditioning input to reconstruct the features. The output will be used 1) as a conditioning input for coding a denser level, and 2) by the \textit{3D Gaussian Generation} module to generate the renderable 3D Gaussians.

\subsubsection{3D Gaussian Generation}

The \textit{3D Gaussian Generation} module architecture shown in Table~\ref{tab:3d_gaussian_generation} is for generating the renderable 3D Gaussians from the reconstructed features. If this module is used at a coarse level (level 8 in our experiments), the 4th layer in Table~\ref{tab:3d_gaussian_generation} will be a 3D transposed convolution, which generates 8 times more points than the input for finer details. If this module is used at a denser level (level 9 in our experiments), the 4th layer in Table~\ref{tab:3d_gaussian_generation} is configured as a Geometry-Invariant 3D convolution, which generates the same number of points as the input. The output includes 14 channels for the 3D Gaussian parameters, \textit{i.e.} $\mu_x, \mu_y, \mu_z, \sigma_x, \sigma_y, \sigma_z, q_w, q_x, q_y, q_z, o, c$.

\subsection{Discussion on Point Cloud Video Compression and Rendering}
\label{sec:discussion}

In this section, we discuss a promising extension of our method to point cloud video compression and rendering. We conduct this experiment with the 8iVFB dataset~\cite{dataset8i}. We compare our method to the standard scheme, whice employs G-PCC for point cloud compression and OpenGL for rendering. Please kindly refer to the supplementary video for a demonstration of results.

As shown, our method is able to achieve a better quality at a lower bit-rate compared to the standard scheme. More importantly, when we look closer to the 3D surface, the holes / gaps between points will be visible in the standard scheme, which degrade the visual quality. In contrast, our method produces 3D Gaussians that approximate the smooth surface. Therefore, it does not suffer from the same issue and shows less visual artifacts.

Nevertheless, since the main focus of this paper is on joint point cloud compression and rendering for one frame, our method does have limitations in point cloud video compression and rendering. For example, we didn't consider the temporal coherence between frames, which is important for both compression and reconstruction. We notice some flickering effects in the rendered video, which is caused by the lack of temporal coherence. We believe that our method can be extended to point cloud video compression and rendering by incorporating temporal consistency loss and temporal modeling designs. For example, we can use warped previous frames as a conditioning input for the \textit{Feature Squeeze} and \textit{Feature Reconstruct} modules to reduce the bit-rates. We can also use this information to improve the temporal coherence in the rendering process, forming spatio-temporal 4D Gaussians. We leave these extensions as a promising future work.

\begin{table*}[h]
  \centering
  \caption{The architecture of the \textit{Feature Convert} module.}
  \scriptsize
  \label{tab:feature_convert}
  \begin{tabular}{c|c|c|c|c|c}
      \toprule
      Layer & Layer Type & Input Channel & Output Channel & Kernel Size & Activation \\
      \midrule
      1 & Geom-Invariant 3D Conv. & $C_\text{in}$ & 64 & $3\times3$ & None \\
      2 & InceptionResBlock & 64 & 64 & - & ReLU \\
      3 & InceptionResBlock & 64 & 64 & - & ReLU \\
      4 & Geom-Invariant 3D Conv. & 64 & 64 & $3\times3$ & None \\
      \bottomrule
  \end{tabular}
\end{table*}

\begin{table*}[h]
    \centering
    \caption{The architecture of the \textit{Feature Squeeze} module.}
    \scriptsize
    \label{tab:feature_squeeze}
    \begin{tabular}{c|c|c|c|c|c}
        \toprule
        Layer & Layer Type & Input Channel & Output Channel & Kernel Size & Activation \\
        \midrule
        1 & Per-point Linear Layer & 64 + 64 & 64 & $1\times1$ & ReLU \\
        2 & Per-point Linear Layer  & 64 & 8 & $1\times1$ & None \\
        \bottomrule
    \end{tabular}
\end{table*}

\begin{table*}[h]
    \centering
    \caption{The architecture of the \textit{Predictive Entropy Model} module.}
    \scriptsize
    \label{tab:entropy_model}
    \begin{tabular}{c|c|c|c|c|c|c}
        \toprule
        Layer & Prev. & Layer Type & Input Channel & Output Channel & Kernel Size & Activation \\
        \midrule
        1 & - & Geom-Invariant 3D Conv. & 64 & 64 & $3\times3$ & ReLU \\
        2 & 1 & \makecell{Geom-Invariant \\InceptionResBlock} & 64 & 64 & - & ReLU \\
        3 & 2 & \makecell{Geom-Invariant \\ InceptionResBlock} & 64 & 64 & - & ReLU \\
        4 & 3 & Per-point Linear Layer & 64 & 8 & $1\times1$ & None \\
        5 & 3 & Per-point Linear Layer & 64 & 8 & $1\times1$ & Exp \\
        \bottomrule
    \end{tabular}
\end{table*}

\begin{table*}
    \centering
    \caption{The architecture of the \textit{Feature Reconstruct} module.}
    \scriptsize
    \label{tab:feature_reconstruct}
    \begin{tabular}{c|c|c|c|c|c}
        \toprule
        Layer & Layer Type & Input Channel & Output Channel & Kernel Size & Activation \\
        \midrule
        1 & Per-point Linear Layer & 8 + 64 & 64 & $1\times1$ & ReLU \\
        2 & InceptionResBlock & 64 & 64 & - & ReLU \\
        3 & InceptionResBlock & 64 & 64 & - & ReLU \\
        4 & Per-point Linear Layer & 64 & 64 & $1\times1$ & None \\
        \bottomrule
    \end{tabular}
\end{table*}

\begin{table*}
    \centering
    \caption{The architecture of the \textit{3D Gaussian
    Generation} module.}
    \scriptsize
    \label{tab:3d_gaussian_generation}
    \begin{tabular}{c|c|c|c|c|c}
        \toprule
        Layer & Layer Type & Input Channel & Output Channel & Kernel Size & Activation \\
        \midrule
        1 & Geom-Invariant Conv. & 64 & 64 & $3\times3$ & ReLU \\
        2 & ResBlock & 64 & 64 & - & ReLU \\
        3 & ResBlock & 64 & 64 & - & ReLU \\
        4 & \makecell{Geom-Invariant Conv. or \\ 3D Transposed Conv.} & 64 & 64 & $2\times2$ & None \\
        5 & ResBlock & 64 & 64 & - & ReLU \\
        6 & ResBlock & 64 & 64 & - & ReLU \\
        7 & Geom-Invariant Conv. & 64 & 14 & $3\times3$ & None \\
        \bottomrule
    \end{tabular}
\end{table*}

\end{document}